# Potential Functions based Sampling Heuristic For Optimal Path Planning*

Ahmed Hussain Qureshi[1,2], Yasar Ayaz[1]

[1]Robotics And Intelligent Systems Engineering (RISE) Lab
Department of Robotics and Artificial Intelligence
School of Mechanical And Manufacturing Engineering (SMME)
National University of Sciences And Technology (NUST)
H-12 Campus, Islamabad, 44000, Pakistan.

[2]Department of System Innovation
Graduate School of Engineering Science
Osaka University, 1-3 Machikaneyama, Toyonaka, Osaka, Japan.

April 2, 2017

## Abstract

Rapidly-exploring Random Tree Star(RRT*) is a recently proposed extension of Rapidly-exploring Random Tree (RRT) algorithm that provides a collision-free, asymptotically optimal path regardless of obstacles geometry in a given environment. However, one of the limitation in the RRT* algorithm is slow convergence to optimal path solution. As a result it consumes high memory as well as time due to the large number of iterations utilised in achieving optimal path solution. To overcome these limitations, we propose the Potential Function Based-RRT* (P-RRT*) that incorporates the Artificial Potential Field Algorithm in RRT*. The proposed algorithm allows a considerable decrease in the number of iterations and thus leads to more efficient memory utilization and an accelerated convergence rate. In order to illustrate the usefulness of the proposed algorithm in terms of space execution and convergence rate, this paper presents rigorous simulation based comparisons between the proposed techniques and RRT* under different environmental conditions. Moreover, both algorithms are also tested and compared under non-holonomic differential constraints.

## 1 Introduction

Motion planning involves collision free navigation of a robot from an initial start region to a goal region in given environments. Applications of this field are not only restricted to robotics [12], but finds application in various other fields such as computer animation [10], medical [23], modern industry [14] and in our daily life [18]. Due to its comprehensive application, many algorithms have been developed in the field of motion planning. Two major classes of motion planning algorithms exist; one is complete algorithms, which successfully return a solution in finite time, if one exists, and reports failure if a feasible solution does not exist. The other class of algorithms does

---

not assure full completeness but does assure probabilistic or resolution completeness. Many complete motion planning algorithms exist [22] [17] but are often computationally inefficient [3] for common practical applications [6]. Algorithms providing resolution completeness include the example of Artificial Potential Fields (APF) [8] and cell decomposition methods [2]. These algorithms, however, are only effective in problem solving if the resolution parameter of the grid is finely tuned. Moreover Artificial potential fields (APF) perform pure exploitation. Exploitation makes the planner greedy as it assumes that the provided information is sufficient for computing a path solution. Although pure exploitation allows APF to quickly compute the solution but it also causes APF to suffer from the problem of local minima [9]. The other resolution complete algorithm i.e., cell decomposition methods involves extremely large numbers of cells which makes it computationally heavy. Therefore, these limitations make these methods unsuited for the motion planning of robots placed in complex environments. To overcome the problem of inefficiency, computationally efficient sampling based algorithms [11] were introduced. Sampling based algorithms perform pure exploration of configuration space so as to improve the planners' understanding of the given space. The most effective of these were Probabilistic Road Maps (PRM) [7] and Rapidly-exploring Random Trees [13], ensuring *probabilistic completeness*. This means that as the number of iterations approaches infinity, the probability of finding path, if one exists, approaches one. PRM's and its variants are multiple-query methods. However, most online motion planning problems can be solved as single-query problems instead [6]. Furthermore, PRM's need prior computing of the roadmap which is not feasible when the environment it is operating in is unknown. Also, computing a road map during run time is computationally expensive. Rapidly Growing Random Tree algorithms were mainly developed for motion planning problems with differential constraints and for single query problems [13]. Recently, an extension of Rapidly-exploring Random Trees algorithm called Rapidly-exploring Random Tree star (RRT*) [6] was proposed which computes an initial path same as RRT but continues to perform further iterations, increasing the number of samples in the configuration space to optimize this initial path, thereby ensuring asymptotic optimality [5]. This feature is not provided by the RRTs [6]. However some major constraints that still exist in RRT*, due to pure exploration, are its slow rate of convergence in determining an optimal path solution and its high memory requirements. Large memory utilization occurs due to the large number of iterations required to find the optimal path. Although computing an optimal path solution is itself a challenging problem, but fast convergence to optimal solution is also important for most online motion planning problems [4].

This paper introduces the idea of potentially guided, directionalized sampling by incorporating Artificial Potential Field Algorithm (APF) [8] into RRT*, thus resulting into guided exploration of given environment. This speeds up convergence towards a solution as directionalized samples reduces the number of iterations, and consequently execution time, required to achieve an optimal path. Artificial Potential Field (APF) algorithms are known for their simplicity and strong mathematical analysis but their applications are limited to a state space of up to five dimensions due to its inability to work in local minima environments [9]. APF primarily uses the effect of unreal forces that act on the robot, generated by both the goal and obstacle regions. This idea of fictitious forces was given by [8]. A similar concept of directionalizing random samples through Artificial potential fields was initially introduced as Potential Guided Directionalized-RRT* (PGD-RRT*) [20]. Although PGD-RRT* finds an initial path very quickly as compared to RRT* but it fails to converge to optimal path solution. Our proposed Potential Functions based RRT* (P-RRT*) is a variant of the previously proposed PGD-RRT* and extension of APGD-RRT* [21]. It efficiently inculcates APF into RRT* to keep the balance between exploitation and exploration i.e., guided exploration of given space. This inculcation helps P-RRT* to direct random samples in the direction of decreasing potential to provide a quick, optimal solution. Moreover, this also results in lesser dispersion of samples in the configuration space and provides a more memory efficient solution operating at a much faster rate compared to RRT*. This idea of guiding samples for improving and theoretically characterizing the convergence rate of asymptotically optimal sampling-based algorithms is novel. As per the authors knowledge, no such technique exists of guiding random samples by APF for fast optimal motion planning. This new algorithm has been evaluated under different scenarios including the local minima environment. It has been observed that in almost all the cases, our P-RRT* is more efficient than RRT*. The remainder of the paper is organized as follows. Section II addresses the problem definition, Section III explain the RRT* and Artificial Potential Fields algorithms while Section IV describes P-RRT* path planning algorithm in detail. Section V gives a brief outline of the implementation of P-RRT* and RRT* under non-holonomic differential constraints. Section VI presents analysis of the of the proposed algorithm in terms of probabilistic completeness, asymptotic optimality, convergence to optimal solution and computational complexity. Section VII provides experimental evidence in support of theoretical results presented in the previous section; whereas Section VIII concludes the paper, also suggesting some future areas of research in this particular domain.

## 2 Problem Definition

This section presents three motion planning problems we will be addressing in this paper, along with the notations we will be using to describe them.

Given a set $Q$, a sequence denoted as $\{q_i\}_{i \in \mathbb{N}}$ in this set is a mapping from $\mathbb{N}$ to $Q$ i.e., $i \in \mathbb{N}$ is mapped to $q_i \in Q$. Moreover, for the algorithms described in this paper, each set $Q$ is equipped with remove and add procedures such that $Q.\text{add}(q) := Q \cup \{q\}$ while $Q.\text{remove}(q) := Q \backslash \{q\}$. Let the constant $\lambda \in \mathbb{R}_+$ be a small step size. Let $X \subset \mathbb{R}^d$ represent the given state space, where $d$ denotes the dimension of state space i.e., $d \in \mathbb{N} : d \geq 2$. The obstacle and obstacle-free state space is defined as $X_{\text{obs}} \subset X$ and $X_{\text{free}} = X \backslash X_{\text{obs}}$ respectively. The initial state is denoted as $x_{\text{init}} \in X_{\text{free}}$ while goal region is denoted as $X_{\text{goal}} \subset X_{\text{free}}$. The procedure $\mu(\cdot)$ [1] Given $x_1, x_2 \in X$, the Euclidean distance between these two states is defined as $d(x_1, x_2) \in \mathbb{R}$. The spherical region centered at any state $x \in X$ of radius $r \in \mathbb{R} | r > 0$ is represented by $\mathfrak{B}_{x,r} := \{y \in X : d(y, x) \leq r\}$. Let $U : \mathbb{R}^d \to \mathbb{R}$ denotes the artificial potential function. The variable $\tau : [0, 1] \to X$ is a path having non-negative and non-zero scalar length. This path $\tau$ is considered *feasible* if it connects $x_{\text{init}}$ and $x \in X_{\text{goal}}$, i.e. $\tau(0) = x_{\text{init}}$ and $\tau(1) \in X_{\text{goal}}$, and lies in the obstacle-free space $X_{\text{free}}$. Problem 1 formalizes the feasibility problem of path planning.

**Problem 1 (Feasible Path Planning)** *Given a triplet $\{X, X_{\text{free}}, X_{\text{obs}}\}$, an initial state $x_{\text{init}}$ and a goal region $X_{\text{goal}} \subset X_{\text{free}}$, find a path $\tau : [0, 1] \to X_{\text{free}}$ such that $\tau(0) = x_{\text{init}}$ and $\tau(1) \in X_{\text{goal}}$.*

Let $\sum_{\text{feasible}}$ denote the set of all feasible trajectories in the obstacle-free configuration space $X_{\text{free}}$. The cost function $c(\cdot)$ finds the path length in terms of Euclidean distance function. Problem 2 formalizes the optimal path planning problem; finding a feasible path with minimum cost $c^*$.

**Problem 2 (Optimal Path Planning)** *Assuming that a solution to problem 1 exists and provided with the set of all feasible trajectories $\sum_{\text{feasible}}$, find a path $\tau^* \in \sum_{\text{feasible}}$ such that $c(\tau^*) = \{\min_{\tau \in \sum_{\text{feasible}}} c(\tau)\}$.*

Let $t \in \mathbb{R}$ denote the time taken by the algorithm to find a set of all feasible paths $\sum_{\text{feasible}}$ (solution to problem 1) and computing the optimal path $\tau^* \in \sum_{\text{feasible}}$. The fast path planning problem formalized in problem statement 3 indicates that this optimal path solution must be determined in least possible time.

**Problem 3 (Fast Path Planning)** *Find the*

---

[1] The procedure $\mu(\cdot)$ provides the Lebesgue measure of any given state space e.g. $\mu(X)$ denotes the Lebesgue measure of the whole state space $X$. Lebesgue measure is also called d-dimensional volume of the given space. provides the Lebesgue measure of any given state space.

*solution to problems 1 and 2, if one exists, in least possible time $t \in \mathbb{R}$.*

## 3 Related Work

This section briefly explains Optimal Rapidly-exploring Random Trees (RRT*) and Artificial Potential Fields (APF) algorithm, which form the basis of our proposed Potential Function Based-RRT* (P-RRT*) algorithm. P-RRT* uses Artificial Potential Fields to guide the random samples picked by RRT* towards the goal for further optimization.

### 3.1 RRT*

This section formally presents the Rapidly-exploring Random Tree Star (RRT*) algorithm [6] that is an extension of the standard RRTs algorithm. Algorithm 1 is slightly modified implementation of RRT*. In this modification, improvements were made to original algorithm in order to enhance computational efficiency of RRT* by reducing the number of calls to its collision checking procedure [19]. Following is a brief description of the main processes involved in its execution:

*Sampling:* The procedure RandomSample($n$) randomly samples the given obstacle-free region $X_{\text{free}}$ to get independent, uniformly distributed configurations.

*Nearby Nodes:* Considering a configuration $x \in X$ and a random tree $T = (V, E)$ where $V \subset X$ and the number of vertices in $V$ is defined as $n := |V|$, the procedure NearbyNodes($T, x, n$) provides a set of nodes $X_{\text{near}} \subset V$ lying within a ball of radius $r$ centered at $x$ i.e.,

$$\text{Nearby}(T, x, n) := \{v \in V : d(v, x) \leq r := \gamma(\frac{logn}{n})^{1/d}\}$$

where $\gamma$ is an independent constant such that $\gamma > \gamma^* := (2(1 + 1/d))^{1/d}\left(\frac{\mu(X_{\text{free}})}{\zeta_{\mathfrak{B}}}\right)^{1/d}$ and $d$ represents the dimension of the configuration space.

*Nearest Node:* Given the configuration $x \in X$, the tree $T = (V, E)$ where $V \subset X$, the NearestNode($x, T$) procedure returns the node $v \in V$ that is nearest to the configuration $x$ in terms of Euclidean distance. This procedure can also be summarize as:

$$\text{NearestNode}(x, T) = \text{argmin}_{v \in V} d(v, x)$$

*Lists and Sorting:* Given the set $X' \subset X$ and a random state $x \in X$, the procedure GetTuple($x, X'$) returns the sorted list $L$. Algorithm 2 provides the pseudocode of this procedure. Each element of this list comprises of cost $c \in \mathbb{R} : c > 0$, state $x' \in X'$ and the path $\tau$. The list $L$ is equipped with add and sort functions, the former works similar to the one for sets while the latter sorts the list $L$ in ascending order of cost.

*Extending:* Given the two states $x_1, x_2 \in X$, the function ExtendTo($x_1, x_2$) returns a path $\tau : [0, 1] \to X$ such that $\tau(0) = x_1$ and $\tau(1) = x_2$. The extension procedure provides the straight path, i.e., $\tau(s) = (1 - s)x_1 + sx_2; \forall s \in [0, 1]$.

*Collision checking:* Given two configurations $x_1, x_2 \in X$, a path $\tau : [0, 1]$ such that $\tau(0) = x_1$ and $\tau(1) = x_2$, the procedure CollisonFree($\tau$) returns true if the path $\tau$ belongs to obstacle-free space $X_{\text{free}}$

---

**Algorithm 1:** RRT*($x_{\text{init}}$)

1 $V \leftarrow \{x_{\text{init}}\}; E \leftarrow \emptyset; T \leftarrow (V, E)$;
2 **for** $n \leftarrow 0$ **to** $N$ **do**
3     $x_{\text{rand}} \leftarrow$ RandomSample($n$);
4     $X_{\text{near}} \leftarrow$ NearbyNodes($T, x_{\text{rand}}, n$);
5     **if** $X_{\text{near}} = \emptyset$ **then**
6        $X_{\text{near}} \leftarrow$ NearestNode($x_{\text{rand}}, T = (V, E)$);
7     $L \leftarrow$ GetTuple($x_{\text{rand}}, X_{\text{near}}$);
8     $x_{\text{parent}} \leftarrow$ SelectBestParent($L$);
9     **if** $x_{\text{parent}} \neq \emptyset$ **then**
10       $T = (V, E) \leftarrow$ InsertNode($x_{\text{rand}}, x_{\text{parent}}, T = (V, E)$);
11       $E \leftarrow$ RewireNodes($x_{\text{rand}}, L, E$);
12 **return** $T = (V, E)$;

**Algorithm 2:** GetTuple($x_{\text{rand}}, X_{\text{near}}$)
1 $L \leftarrow \emptyset$;
2 **for** $x' \in X_{\text{near}}$ **do**
3 $\quad \tau \leftarrow \text{ExtendTo}(x', x_{\text{rand}})$;
4 $\quad c \leftarrow c(x') + c(\tau)$;
5 $\quad L \leftarrow (x', c, \tau)$;
6 $L.\text{sort}()$;
7 **return** $L$;

otherwise it reports failure.

Algorithm 1 explains the RRT* algorithm. Once initialized, the RRT* algorithm begins its iterative processing by picking random samples, $x_{\text{rand}}$, from the obstacle-free configuration space $X_{\text{free}}$ (Line 3). The algorithm then determines the set of near vertices $X_{\text{near}}$, described as the vertices of the random tree that lie within the ball region centered at $x_{\text{rand}}$. If no such vertices exist and the set $X_{\text{near}}$ computed by the NearbyNodes procedure is empty, the set $X_{\text{near}}$ is then filled by the NearestNode function (Line 4-6). Once populated, the set $X_{\text{near}}$ is sorted, forming a tuple arranged in ascending order of cost (Line 7). The sorted list $L$ is used by the SelectBestParent function ( Line 8), which returns the best parent vertex $x_{\text{parent}} \in X_{\text{near}}$ through which the point $x_{\text{init}}$ and $x_{\text{rand}}$ can be connected in obstacle free configuration space. Algorithm 3 outlines the implementation SelectBestParent procedure which iterates through each element in the sorted list $L$ and terminates by returning the vertex through which $x_{\text{rand}}$ can be connected to the tree in obstacle-free space. Once the algorithm finds such a state, i.e, the best parent vertex $x_{\text{parent}}$ gets filled, $x_{\text{parent}}$ is added to the tree by making $x_{\text{rand}}$ its child and then rewiring the random tree (Line 9-11). Algorithm 4 gives the pseudocode of this rewiring process. RRT* examines each vertex $x'$ in list $L$. If the cost of a path lying in obstacle free space and connecting the initial point $x_{\text{init}}$ to $x'$ through the random sample $x_{\text{rand}}$ is less than the existing cost of reaching $x'$ (Algorithm 4 Line 1-3), then $x_{\text{rand}}$ is made into the parent of $x'$ (Algorithm 4 Line 4-5). Otherwise, no change is made to the tree and RRT* moves on to examine another vertex. This process is performed iteratively for each vertex $x'$ in the sorted list $L$.

**Algorithm 3:** SelectBestParent($L$)
1 **for** $(x', c, \tau) \in L$ **do**
2 $\quad$ **if** CollisionFree($\tau$) **then**
3 $\quad\quad$ **return** $x'$;
4 **return** $\emptyset$;

**Algorithm 4:** RewireNodes($x_{\text{rand}}, L, E$)
1 **for** $(x', c, \tau) \in L$ **do**
2 $\quad$ **if** $\big(c(x_{\text{rand}}) + c(\tau)\big) < c(x')$ **then**
3 $\quad\quad$ **if** CollisionFree($\tau$) **then**
4 $\quad\quad\quad$ $x'_{\text{parent}} \leftarrow \text{GetParent}(E, x')$;
5 $\quad\quad\quad$ $E.\text{remove}(x'_{\text{parent}}, x')$;
6 $\quad\quad\quad$ $E.\text{add}(x_{\text{rand}}, x')$;
7 **return** $E$;

### 3.2 Artificial Potential Fields

APF by [8] utilizes gradient descent planning that tries to minimize artificial potential energy. The main robot, denoted as $x \in X$, and the goal region $X_{\text{goal}}$ is assigned an attractive potential $U_{\text{att}}$ while obstacle regions are assigned repulsive potentials $U_{\text{rep}}$. This causes the robot $x$ to be attracted towards the goal and repelled by the obstacles. These attractive and repulsive potentials cause the robot to experience a force $\overrightarrow{F}$ equal to the negated gradient of potentials i.e., $\overrightarrow{F} = -\bigtriangledown U$. Under the influence of both attractive and repulsive forces, the robot moves down the slop and reaches the goal region safely i.e., without any collisions. The constants $K_a$ and $K_r$ indicate the scaling factors that are used to scale the magnitude of attractive and repulsive potential, respectively. These factors are dependent upon the configuration space. Attractive potential experienced by the robot is formulated in equation 1. It varies quadratically when the distance function $d(x, x_\text{g}) > d_\text{g}^*$. The parameter $d_\text{g}^*$ is the radius of the circular boundary centered at the goal state $x_\text{g} \in X_\text{goal}$, defining the quadratic range. This quadratic function allows the robot to quickly move towards the goal region due to high attractive forces created between the robot at position $x$ and the goal state $x_\text{g} \in X_\text{goal}$. However, once the robot enters the circular region centered around $x_g$, the attractive potential starts to vary conically. This allows the robot to move slowly when it comes close to the goal due to reduced attractive potential, thereby preventing it from overshooting the goal region. The attractive force is formulated in equation 2.

$$U_\text{att} = \begin{cases} K_a d^2(x, x_\text{g}) & d(x, x_\text{g}) > d_\text{g}^* \\ K_a(d_\text{g}^* d(x, x_\text{g}) - (d_\text{g}^*)^2) & d(x, x_\text{g}) \leq d_\text{g}^* \end{cases} \quad (1)$$

$$\overrightarrow{F}_\text{att} = \begin{cases} -2K_a d(x, x_\text{g}) & d(x, x_\text{g}) > d_\text{g}^* \\ -2d_\text{g}^* K_a \dfrac{x - x_\text{g}}{d(x, x_\text{g})} & d(x, x_\text{g}) \leq d_\text{g}^* \end{cases} \quad (2)$$

Repulsive potential generated by the obstacles $X_\text{obs} \subset X$ is formulated in equation 4. Equation (3) is used to calculate the distance $d_\text{min}$ of the robot $x$ from the closest vertex in the obstacle space $X_\text{obs}$. Repulsive potential is considered zero if the distance $d_\text{min}$ is greater than a constant value $d_\text{obs}^*$. Such a situation indicates that the robot is at a large distance from the nearest obstacle region. Therefore, to allow the robot to move quickly towards the goal, the repulsive potential is made zero as indicated in equation 4.

$$d_\text{min} = \min_{x' \in X_\text{obs}} d(x, x') \quad (3)$$

$$U_\text{rep} = \begin{cases} \dfrac{1}{2} K_r \left( \dfrac{1}{d_\text{min}} - \dfrac{1}{d_\text{obs}^*} \right)^2 & d_\text{min} \leq d_\text{obs}^* \\ 0 & d_\text{min} > d_\text{obs}^* \end{cases} \quad (4)$$

The repulsive force generated due to $X_\text{obs}$ is presented in equation 6 and is equal to the negated gradient of repulsive potential indicated in equation 4. The negated gradient of equation 3 is formulated as equation 5, where $x'$ is the nearest obstacle state in the obstacle space i.e., $x' \in X_\text{obs}$ from the robot's current position $x \in X$.

$$\frac{\partial d_\text{min}}{\partial x} = \frac{(x - x')}{d(x, x')} \quad (5)$$

$$\overrightarrow{F}_\text{rep} = \begin{cases} K_r(\dfrac{1}{d_\text{obs}^*} - \dfrac{1}{d_\text{min}})\dfrac{1}{d_\text{min}^2}\dfrac{\partial d_\text{min}}{\partial x} & d_\text{min} \leq d_\text{obs}^* \\ 0 & d_\text{min} > d_\text{obs}^* \end{cases} \quad (6)$$

The net overall potential $U$ is the sum of both attractive and repulsive potentials, while the global force $\overrightarrow{F}$ can be formalized as $\overrightarrow{F} = -\bigtriangledown U$. Algorithm 5 indicates the gradient descent procedure used in Artificial Potential Fields where $\lambda$ is a small incremental distance. The algorithms keeps on iterating until the robot reaches the configuration having zero potential energy (Line 2). However, a configuration where the potential energy is zero can indicate two things; either the robot has reached the goal region or it is stuck in the local minima configuration.

---

**Algorithm 5:** GradientDescent($x_\text{init}$)

1: $x \leftarrow x_\text{init}$;
2: **while** $\bigtriangledown U \neq 0$ **do**
3: $\quad \overrightarrow{F} \leftarrow$ PotentialGradient($x$);
4: $\quad x \leftarrow x + \lambda(\dfrac{\overrightarrow{F}}{|\overrightarrow{F}|})$;
5: **end while**

---

## 4 P-RRT*

In this section, we present an extension of RRT* called Potential Function Based-RRT* (P-RRT*), which incorporates the Artificial Potential Field [15] algorithm into RRT*. Further explanations are given in the discussion below.

Let a potentially guided, random sample be defined as $x_\text{prand} \in X_\text{free}$. The random state $x_\text{rand} \in X_\text{free}$ is incrementally directed downhill in the direction of decreasing attractive potential field gradient by a small discrete step denoted as $\lambda \in \mathbb{R}_+$. Attractive potential gradient decreases as the random sample approaches closer to the goal region. Algorithm 6 outlines the implementation of P-RRT* algorithm, in this there is only one additional procedure i.e., RGD($x$) which is executed just after the sampling procedure. The random sample $x_{\text{rand}}$ is augmented with attractive potential field to get an improved sample $x_{\text{prand}}$, and from now on the algorithm treats $x_{\text{prand}}$ as its random sample as shown in the algorithm 6. P-RRT*, uses Randomized Gradient Descent Planning for computing $x_{\text{prand}} \in X_{\text{free}}$. Gradient Descent planning explained in the previous section computes the next state as a function of the previous state and works iteratively until $|\nabla U| \to 0$, as shown in Algorithm 5. However, in Randomized Gradient Descent (RGD) Planning, next state is independent of the previous state and for each iteration, a random sample $x_{\text{rand}} \in X_{\text{free}}$ is seeded into the RGD($x_{\text{rand}}$). This random sample is then moved incrementally in the direction of decreasing potential by step size $\lambda$ to generate $x_{\text{prand}} \in X_{\text{free}}$. It should be noted here that the constant $\lambda$ is a small incremental step size as stated in section 2.

**Algorithm 6:** P-RRT*($x_{\text{init}}$)
---
1   $V \leftarrow \{x_{\text{init}}\}; E \leftarrow \emptyset; T \leftarrow (V, E)$;
2   **for** $n \leftarrow 0$ **to** $N$ **do**
3      $x_{\text{rand}} \leftarrow \text{RandomSample}(n)$;
4      $x_{\text{prand}} \leftarrow \text{RGD}(x_{\text{rand}})$;
5      $X_{\text{near}} \leftarrow \text{NearbyNodes}(T, x_{\text{prand}}, n)$;
6      **if** $X_{\text{near}} = \emptyset$ **then**
7         $X_{\text{near}} \leftarrow \text{NearestNode}(x_{\text{prand}}, T = (V, E))$;
8      $L \leftarrow \text{GetTuple}(x_{\text{prand}}, X_{\text{near}})$;
9      $x_{\text{parent}} \leftarrow \text{SelectBestParent}(L)$;
10     **if** $x_{\text{parent}} \neq \emptyset$ **then**
11        $T = (V, E) \leftarrow \text{InsertNode}(x_{\text{prand}}, x_{\text{parent}}, T = (V, E))$;
12        $E \leftarrow \text{RewireNodes}(x_{\text{prand}}, L, E)$;
13   **return** $T = (V, E)$;

Since, all the procedure used by P-RRT* are same except RGD($x$), therefore only RGD($x$) procedure is explained here. Following are the set of procedures on which the RGD($x_{\text{rand}}$) function relies.

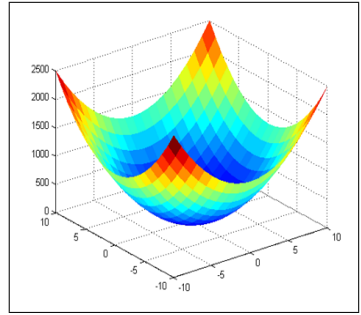

Figure 1: Attractive potential field experienced by random sample. Vertical axis corresponds to magnitude of potential and horizontal axis corresponds to position.

*Attractive Potential Gradient (APG):* The proposed algorithm only utilizes quadratic variation in the attractive potential fields, instead of shifting between conical and quadratic variation as done in the original Artificial Potential Fields algorithm. In APF, the robot itself is considered under the influence of potential fields. Therefore, conical attractive potential is used to avoid the robot overshooting the goal region. Since in our proposed algorithm, it is the random samples that are under the influence of potential fields, overshooting is not an issue in our case. Hence, the need for conical variation of potential is eliminated. This quadratic attractive field, shown in figure 1, is described by Equations 7 and 8. In figure 1, the horizontal axis is the position of the random sample $x_{\text{rand}}$, where the goal region $X_{\text{goal}}$ is at its origin $(0, 0)$. The vertical axis corresponds to the quantity of attractive potential field present. This well shaped curve indicates that farther the random sample $x_{\text{rand}} \in X_{\text{free}}$ from the origin or goal, greater will be the attractive field.

$$U_{\text{att}} = d^2(x_{\text{rand}}, x_{\text{goal}}) : x_{\text{goal}} \in X_{\text{goal}} \qquad (7)$$

$$\overrightarrow{F}_{\text{att}} = -2d(x_{\text{rand}}, x_{\text{goal}}) : x_{\text{goal}} \in X_{\text{goal}} \qquad (8)$$

*Nearest Obstacle Configuration:* This procedure computes the nearest obstacle configuration from the random sample $x_{\text{rand}}$. This procedure utilizes equation (3), however in this case, the variable $x$ (i.e., robot configuration) in equation (3) is replaced by the variable $x_{\text{rand}}$ (i.e., the random sample). It should be noted that proposed procedure only computes $d_{\text{min}}$ from the point in the obstacle space that is nearest to the random sample. This signifies that just like RRT*, P-RRT* does not require information about obstacle geometry.

Algorithm 7 outlines the working of the RGD($x_{\text{rand}}$) procedure. This function first computes the quadratic attractive potential gradient acting on the independent and identically distributed (iid) sample $x_{\text{rand}}$ (Line 3) and then it computes the distance $d_{\text{min}}$ of the random sample from the nearest obstacle configuration (Line 4). Under the influence of the attractive field (Line 8), this random sample is directed in small incremental steps $\lambda \in \mathbb{R}_+$ towards the goal. If at any point, $d_{\text{min}} \leq d_{\text{obs}}^*$, the procedure terminates immediately, returning the new directed sample $x_{\text{prand}}$, otherwise, the procedure continues to direct the random sample for a limited number of iterations $k \in \mathbb{N}$ and then self terminates. The constant $d_{\text{obs}}^*$ represents the distance from the obstacle space and it is kept very small so that the directed random sample $x_{\text{rand}}$ is allowed to move very close to the obstacle region $X_{\text{obs}}$. The significance of keeping $d_{\text{obs}}^*$ small will be discussed later in the analysis section. Moreover, the value of $k$ is chosen so as to maintain the balance between exploitation and exploration. Large value of $k$ will result in more exploitation of the configuration space than exploration. Similarly, a very small value will result in more exploration than exploitation.

## 5 Implementation using non-holonomic wheeled mobile robot (WMR) Poineer 3-DX

In this section, a brief outline of the implementation of P-RRT* and RRT* using non-holonomic differential drive Poineer 3-DX robot is presented. Since discussion on kinematic and dynamic model (Kinodynamic model) of Poineer 3-DX does not come

---

**Algorithm 7:** RGD($x_{\text{rand}}$)

1   $x_{\text{prand}} \leftarrow x_{\text{rand}}$;
2   **for** $n \leftarrow 0$ **to** $k$ **do**
3     $\overrightarrow{F}_{\text{att}} \leftarrow \text{APG}(X_{\text{goal}}, x_{\text{prand}})$;
4     $d_{\text{min}} \leftarrow \text{NearestObstacle}(X_{\text{obs}}, x_{\text{prand}})$;
5     **if** $d_{\text{min}} \leq d_{\text{obs}}^*$ **then**
6       **return** $x_{\text{prand}}$;
7     **else**
8       $x_{\text{prand}} \leftarrow x_{\text{prand}} + \lambda \left( \dfrac{\overrightarrow{F}_{\text{att}}}{|\overrightarrow{F}_{\text{att}}|} \right)$;
9   **return** $x_{\text{prand}}$;

---

under the scope of this paper, therefore detailed description can be seen in [14]. Furthermore the solutions to problems 1-3 are now computed under following non-holonomic constraint of differential drive robot, where $\theta \in \mathbb{R}$, denotes the robot orientation.

$$sin\theta.dx - cos\theta.dy = 0$$

A random configuration $x_{\text{rand}}$ is sampled from the obstacle free space and it is directed towards the goal region (in case of P-RRT* only). The set of near nodes $X_{\text{near}}$ is computed (see Algorithm 1 and 6). Then each node in the $X_{\text{near}}$ set is considered to be the current robot state and a set of allowed control inputs is applied to the robots' Kinodynamic model in order to estimate its future possible state while extending towards the random sample. Results of this estimation are then used for collision checking. The nearest node $x' \in X_{\text{near}}$ which ensures collision-free extension is selected as the best parent for random sample. Moreover, it should also be noted that this procedure of extension is always repeated whenever a tree attempts to connect any two states. For estimation purposes forth order Runge-Kutta method is used and for collision detection tight fitting axis-aligned bounding boxes are employed [16].

# 6 Analysis

## 6.1 Probabilistic Completeness

Most sampling-based algorithms ensure probabilistic completeness. Let $\mathcal{V}_n^{\text{AL}}$ denote the vertices of the tree generated by an algorithm AL after $n$ iterations. Definition 1 formalizes the notion of *probabilistic completeness*, an algorithm is probabilistically complete if the probability of finding a feasible path (solution to problem 1), if one exists, approaches one as the number of iterations approaches infinity.

**Definition 1 (Probabilistic Completeness)**
*Given the path planning problem $\{\mathcal{X}_{\text{free}}, x_{\text{init}}, \mathcal{X}_{\text{goal}}\}$, an algorithm AL ensures probabilistic completeness if and only if $\lim_{n\to\infty} \mathbb{P}(\mathcal{V}_n^{\text{AL}} \cap X_{\text{goal}} \neq \emptyset) = 1$; and the algorithm AL also connects $x_{\text{init}}$ to $x_{\text{goal}} \in X_{\text{goal}}$.*

RRT ensures *probabilistic completeness* and it has also been proved that its variant, i.e. RRT* [6], also inherits this property from the original RRT as formulated in Theorem 1.

**Theorem 1** [6] *Given the path planning problem $\{\mathcal{X}_{\text{free}}, x_{\text{init}}, \mathcal{X}_{\text{goal}}\}$, the probability of finding the solution to Problem 1, if one exists, approaches one as the number of iterations approach infinity, i.e.,*

$$\lim_{n\to\infty} \mathbb{P}(\{\mathcal{V}_n^{\text{RRT}^*} \cap X_{\text{goal}} \neq \emptyset\}) = 1$$

Similar to RRT*, we claim that Theorem 1 holds for P-RRT* as well, which is stated formally in Theorem 2 as follow.

**Theorem 2** *Given a path planning problem, if a feasible path solution exists, then*

$$\lim_{n\to\infty} \mathbb{P}(\{\mathcal{V}_n^{\text{P-RRT}^*} \cap X_{\text{goal}} \neq \emptyset\}) = 1$$

**Sketch of proof:** The proof of above theorem is based on three arguments: 1) By convention we have defined $\mathcal{V}_0^{\text{RRT}^*} = \mathcal{V}_0^{\text{P-RRT}^*} = x_{\text{init}}$ (See Algorithm 1 and 6). Therefore, just like RRT* the random tree generated by P-RRT* necessarily includes $x_{\text{init}}$ as one of its states; 2) Just like RRT*, the tree generated by P-RRT* is also a connected tree i.e., whenever a random sample is chosen, it is connected to its nearest neighbor state within the tree; and 3) P-RRT* directs the random samples towards the goal region $X_{\text{goal}}$, therefore, the probability that the tree generated by P-RRT* will find a goal region approaches to one as the number of iterations approach infinity. Based on the above three arguments, it can be stated that given the path planning problem, the probability that P-RRT* will find a feasible path solution, if one exist, approaches to one as the iterations approaches to infinity. Hence, just like RRT*, the P-RRT* algorithm also ensures *probabilistic completeness*.

Rest of this section is devoted to emphasize one of the important feature of P-RRT* due to goal directed sampling. Let an *attraction sequence* $\mathcal{A} = \{A_0, A_1, \ldots, A_k\}$ of length $k \in \mathbb{R}_+$, be a finite sequence of sets such that (i) $A_0 = \{x_{\text{init}}\}$, (ii) $A_k \in X_{goal}$, and (iii) for each attractor $A_n$, there exists a set called *basin of attraction* $B_n \subseteq X$ such that $d(x,y) < d(x,z)$ for any $x \in A_{n-1}$, $y \in A_n$ and $z \in X \backslash B_n$. Given the attraction sequence of finite length $k$, let $p$ be defined as:

$$\min\left(\frac{\mu(\mathcal{A})}{\mu(X_{\text{free}})}\right); \forall n \in (0, k]$$

For RRT* algorithm, it has been proven that, if there exists a feasible path, then the probability that RRT* fails to find a solution exponentially decays to zero as the number of iterations approach infinity. This is formally stated in theorem 2.

**Theorem 3** [6] *Given a path planning problem $\{\mathcal{X}_{\text{free}}, x_{\text{init}}, \mathcal{X}_{\text{goal}}\}$, if there exits an attraction sequence $\mathcal{A}$ of length $k$, then $\mathbb{P}(\{\mathcal{V}_n^{\text{RRT}^*} \cap \mathcal{X}_{\text{goal}} = \emptyset\}) \leq e^{\frac{-1}{2}(np-2k)}$.*

An *attraction sequence* corresponds to the sequence to which the system eventually evolves. In this case it is the feasible path solution which P-RRT* is aiming to determine. Since P-RRT* directs the random samples toward the goal region, therefore, it can be stated that if there exist a feasible path then the probability that P-RRT* fails to find a solution decays exponentially to zero more quickly as compared to RRT*, as the number of iterations approach infinity. Theorem 4 formally states the above statement.

**Theorem 4** *Given a path planning problem, if a feasible path solution and an attraction sequence of length k exists, then $\lim_{n\to\infty} \mathbb{P}(\{\mathcal{V}_n^{P-RRT^*} \cap \mathcal{X}_{goal} = \emptyset\}) \leq e^{\frac{-1}{2}\alpha(np-2k)}$, where $\alpha \in \mathbb{R}_+$.*

Hence the positive consequence of goal directed sampling by P-RRT* is formalized as follow in theorem 5. Theorem 5 states that if there exist a feasible path, the probability that P-RRT* fails to find a solution decays exponentially faster as compared to RRT*, as the number of iterations approach infinity.

**Theorem 5** *Given a path planning problem, if there exists an attraction sequence of length k exists, then $\lim_{n\to\infty} \mathbb{P}(\{\mathcal{V}_n^{P-RRT^*} \cap \mathcal{X}_{goal} = \emptyset\}) < \mathbb{P}(\{\mathcal{V}_n^{RRT^*} \cap \mathcal{X}_{goal} = \emptyset\}).*

## 6.2 Asymptotic Optimality

The proposed algorithm P-RRT* inherits the *asymptotic optimality* property from the original RRT*. An algorithm is *asymptotically optimal* if it computes a minimum cost continuous path solution $\tau^* : [0,1]$ such that $\tau^*(0) = x_{init}$ and $\tau^*(1) \in X_{goal}$, all the while avoiding any collisions in a cluttered environment. This section analyses the P-RRT* algorithm for its ability to solve problem 2 by ensuring almost sure-convergence to optimal path solution, similar to RRT* based on the assumptions stated below.

**Assumption 1 (Additivity of the cost procedure)**
*For any set of paths in an collision-free space $X_{free}$ i.e. $\tau_1, \tau_2 \in \sum_{feasible}$, the cost function $c(\cdot)$ must satisfy: $c(\tau_1) \leq c(\tau_1|\tau_2) : c(\tau_1|\tau_2) = c(\tau_1) + c(\tau_2)$.*

**Assumption 2 (Continuity of the cost procedure)**
*The procedure $c(\cdot)$ is a uniformly continuous function such that there exists a Lipschitz constant $\varepsilon$ for any two paths $\tau_1 : [0, s_1]$ and $\tau_2 : [0, s_2]$, of similar path lengths i.e., $|c(\tau_1) - c(\tau_2)| \leq \varepsilon \sup_{\psi:[0,1]} \|c(\tau_1(\psi s_1)) - c(\tau_2(\psi s_2))\|$.*

**Assumption 3 ($\delta$-spacing of the obstacle)**
*For any state $x \in X_{free}$, there exists a ball region that lies entirely in collision-free space $X_{free}$ (i.e., $\mathfrak{B}_{x',\delta} \subset X_{free}$) of radius $\delta \in \mathbb{R}_{>0}$ centered around another point $x' \in X_{free}$, such that $x \in \mathfrak{B}_{x',\delta}$.*

Assumption 1 simply states that the longer path has a higher cost than the shorter one. Assumption 2 ensures that two paths of approximately same length that are very close to one another have a similar cost. Finally, Assumption 3 asserts that there exists some collision-free space around trajectories so that the algorithm can converge them to an optimal path solution. Based on the aforementioned assumptions, the following theorem formalizes the *asymptotic optimality* of RRT* algorithm.

**Theorem 6 (Asymptotic optimality of RRT* [6]** *Let Assumptions 1, 2 and 3 hold; then the RRT* algorithm is asymptotic optimal whenever $d \geq 2$ and $\gamma > \gamma^* := 2^d(1 + 1/d)\mu(X_{free})$.*

Recall that the P-RRT* only introduces intelligent sampling heuristic into RRT* thus directionalizing the random samples. The rest of the procedures are same as the RRT*. Therefore theorem 6 holds for P-RRT* just as it holds for RRT* and its proof is similar to the proof of lemma 71 for RRT* [6], based on the Borel-Cantelli lemma. Moreover the value of $\gamma$ is always chosen to be larger than $\gamma^*$ (see definition of procedure NearbyNodes) which ensures that at large number of iterations, the ball of radius $\delta$ centered at $x' \in X_{free}$ will include atleast one node from the tree $T = (V, E)$. This implies that there is a high probability that the rewiring procedure explained earlier will rewire some paths to minimize their cost functions. Hence given two paths $\tau_1, \tau_2 \in \sum_{feasible}$ such that both are the closest to each other in term of path variation $\|\tau_1 - \tau_2\|$, the probability of minimizing path variation to zero is one, when the number of iterations approach infinity. Hence theorem 7 formally states the asymptotic optimality property of P-RRT*.

**Theorem 7 (Asymptotic optimality of P-RRT*)**
Let Assumptions 1, 2 and 3 hold; then the P-RRT* algorithm is asymptotic optimal whenever $d \geq 2$ and $\gamma > \gamma^* := 2^d(1 + 1/d)\mu(X_{\text{free}})$.

## 6.3 Fast convergence to optimal path solution

P-RRT* inherits asymptotic optimality property from the original RRT* algorithm as discussed in the previous section. This section analyses the P-RRT* algorithm for its ability to solve problem 3 by ensuring almost fast convergence to optimal path solution. To understand the notion of fast convergence, a few new terms are introduced which are as follow. Let $\delta \in \mathbb{R}_+$, then any random configuration $x \in X_{\text{free}}$ can be defined as a $\delta$-interior state $X_{\text{int}_\delta}$ if the ball region of radius $\delta$ centered at $x$ lies entirely in an obstacle-free space. Moreover any random configuration can be defined as a $\delta$-exterior state if the ball region of radius $\delta$ centered at $x$ lies partially in an obstacle-free space. Let $X_{\text{int}_\delta}$ and $X_{\text{ext}_\delta}$ be subsets of the obstacle-free space $X_{\text{free}}$. $X_{\text{int}_\delta}$ comprises of all $\delta$-interior states i.e. $X_{\text{int}_\delta} := \{x \in X_{\text{free}} : \mathfrak{B}_{x,\delta} \subseteq X_{\text{free}}\}$, while $X_{\text{ext}_\delta} = X_{\text{free}} \backslash X_{\text{int}_\delta}$. Therefore, $X_{\text{ext}_\delta}$ states are those states that lie close to obstacle region but not inside it. Based upon the aforementioned assumptions, following definitions describe path solutions with *strong $\delta$-clearance* and *weak $\delta$-clearance*, while definition 4 describes the optimal path.

**Definition 2 (Path with strong $\delta$-clearance)** A feasible path solution $\tau : [0, 1]$ is said to have strong $\delta$-clearance if $\tau(s) \in X_{\text{int}_\delta}; \forall s \in [0, 1]$.

**Definition 3 (Path with weak $\delta$-clearance)** A collision-free path $\tau_1 : [0, 1]$ is said to have weak $\delta$-clearance if there exits a collision-free path $\tau_2$ having strong $\delta$-clearance such that i) both paths have same ends i.e. $\tau_1(0) = \tau_2(0)$ and $\tau_1(1) = \tau_2(1)$; ii) path $\tau_1$ can be deformed to $\tau_2$ by a homotopy function $h : [0, 1]$ where $h(0) = \tau_1$, $h(1) = \tau_2$ and $h(s) \mapsto X_{\text{free}}, \forall s \in [0, 1]$ iii) for a range $y \in (0, 1]$, there exists $\delta_y \in (0, \delta]$ such that the homotopy function $h(y)$ has strong $\delta_y$-clearance.

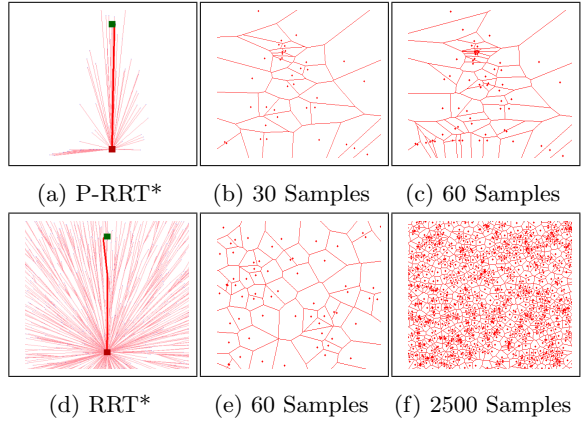

| (a) P-RRT* | (b) 30 Samples | (c) 60 Samples |
| (d) RRT* | (e) 60 Samples | (f) 2500 Samples |

Figure 2: The P-RRT* and RRT* contains a Voronoi bias which causes goal directed and uniform exploration, respectively.

**Definition 4 (Optimal path solution)** A collision-free path is said to be optimal $\tau^*$ if it has weak $\delta$-clearance

The proposed P-RRT* algorithm is build upon the definition of optimal path solution. Since optimal path solution exists in region with *weak $\delta$-clearance*, following lemma states that the proposed sampling heuristic (see Algorithm 7) tries to direct the random samples towards the regions where probability of having optimal path solution is high. While lemma 2 states that P-RRT* directs the random sample towards the goal region. Based upon these two lemmas a theorem 8 has been stated which act as an evidence that P-RRT* ensures rapid convergence to optimal solution.

**Lemma 1** *Given a problem $\{\mathcal{X}_{\text{free}}, x_{\text{init}}, \mathcal{X}_{\text{goal}}\}$,*

*the potential guided sampling heuristic* RGD($x$) *attempts to direct the random sample $x \in \mathcal{X}_{\text{free}}$ so that $\mathbb{P}(x \in \mathcal{X}_{\text{ext}_\delta}) > 0$, for some value of $\delta > 0$.*

**Sketch of proof:** In a cluttered configuration space $X$, the potential guided sampling heuristic directs the random sample $x_{\text{rand}} \in X_{\text{free}}$ down the slope under the influence of the attractive force. This continues until the sample reaches very close to the obstacles space $X_{\text{obs}}$ or the loop limit is reached (see implementation of algorithm 7). Therefore the proposed heuristic always tries to achieve a *weak $\delta$-clearance* for the directed samples. Hence it can be concluded that there exists a good probability that a sample $x_{\text{rand}}$ belongs to the region with *weak $\delta$-clearance* i.e. $X_{\text{ext}_\delta}$. It should also be noted that a large value of $d^*_{\text{obs}}$ will not allow the random sample to reach the region with *weak $\delta$-clearance*, therefore, solution to problem 2 cannot be determined. Moreover *weak $\delta$-clearance* does not require the nodes to be atleast $\delta$ distance away from obstacles. In fact a robustly optimal path with many nodes lying on the boundary of obstacles can still have *weak $\delta$-clearance*. Therefore a very small value of $d_{\text{obs}}$ is needed to solve problem 2.

Following lemma 2 states that the proposed algorithm also guides the random samples towards the goal region.

**Lemma 2** *Given the path planning problem $\{\mathcal{X}_{\text{free}}, x_{\text{init}}, \mathcal{X}_{\text{goal}}\}$, the proposed potential guided sampling heuristic* RGD($x$) *directs the random sample down the potential gradient slope i.e. towards the goal region $X_{\text{goal}}$ in the obstacle-free space $X_{\text{free}}$.*

**Sketch of proof:** An argument for lemma 2 can be given by considering Voronoi regions of the directed nodes as shown in figure 2. Figure 2a depicts the position of starting and goal regions while figure 2b and figure 2c represent the Voronoi diagrams of the directed vertices in the same environment. Unlike RRT* or RRT, the sampling by P-RRT* results in incremental reduction in the size of the Voronoi regions in the direction towards the goal. Therefore it can be said that the proposed heuristic has a Voronoi bias that effectively guides the random samples towards the goal region. Figure 2d to 2f, depicts the Voronoi biasing of RRT*. Moreover, due to this uniform biasing of RRT*, it is able to compute path after 2500 samples, however, due to goal directed Voronoi bias the proposed P-RRT* is able to compute the optimal solution after picking just 60 samples from the configuration space.

Based on lemma 1 and 2 stated above, the distinguishing features of P-RRT* can be highlighted. Without further argument it can be said, as formalized in the following theorem, that our proposed heuristic guides the random samples toward the goal region in such a manner so that the guided samples $x_{\text{prand}}$ also have *weak $\delta$-clearance* in cluttered environments.

**Theorem 8 (Potential guided sampling heuristic RGD(x))** *Let lemma 1 and 2 hold; then the* RGD($x$) *heuristic guides the random samples towards the goal region in such a manner so that $\mathbb{P}(x_{\text{prand}} \in X_{\text{ext}_\delta}) > 0$.*

Hence, based on definition 4 of optimal path solution and theorem 8, it can be concluded that the proposed algorithm P-RRT* computes the optimal path very quickly.

## 6.4 Computational Complexity

This section aims to compare the computational complexities of P-RRT* and RRT*. Let $\mathcal{S}_n^{\text{RRT}^*}$ and $\mathcal{S}_n^{\text{P-RRT}*}$ denote the number of steps executed per iteration $n$ by RRT* and P-RRT* respectively. In the proposed algorithm, the only additional procedure is RGD($x_{\text{init}}$), while the rest of the procedures are exactly the same as used by RRT*. Therefore only the RGD($x_{\text{init}}$) procedure is analyzed for its computational load.

It is to be noted that the procedures in function RGD can be executed in a constant number of steps and are independent of the number of nodes present in the tree. Furthermore, the algorithm has to compute the nearest obstacle configuration from any random state $x \in X_{\text{free}}$ (Equation 3). Finding the nearest neighbor is a well known problem and various algorithms have been implemented in this domain. However, the lower bound of complexity of the algorithm by [1] indicates that nearest neighbor searching requires atleast logarithmic time $\log(n)$. Implementing the computationally optimal algorithm in [1] for computing nearest obstacle configuration under fixed dimension, implies that it has to run in $\Omega(logn)$ time. Since, the algorithm is computationally optimal in fixed dimensions, following lemma states that the computational complexity of executing Equation 3 is nothing more than $\Omega(logn)$ time.

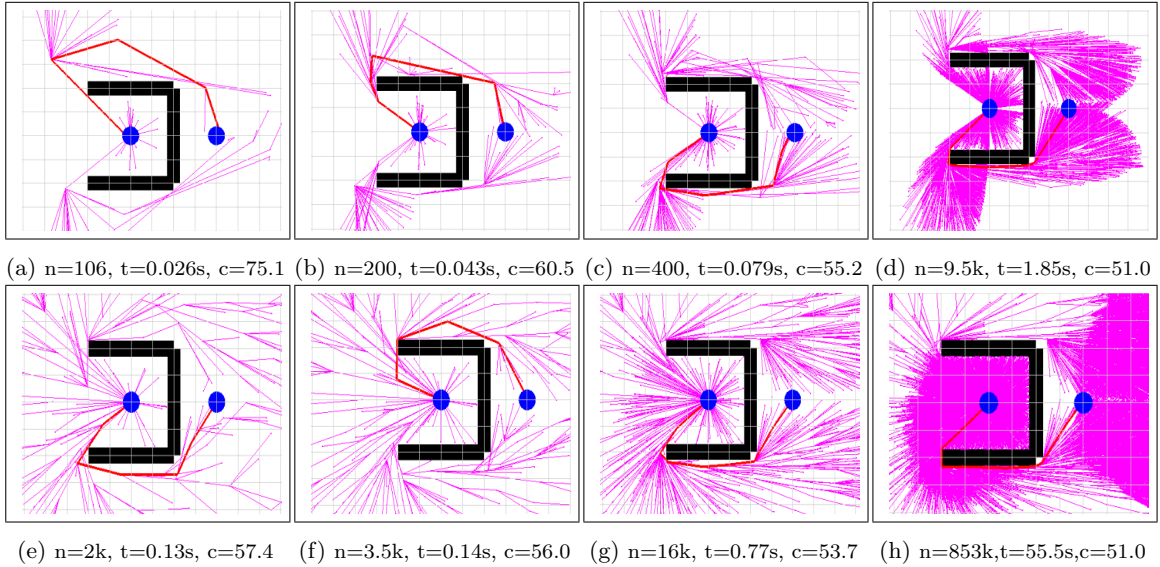

(a) n=106, t=0.026s, c=75.1  (b) n=200, t=0.043s, c=60.5  (c) n=400, t=0.079s, c=55.2  (d) n=9.5k, t=1.85s, c=51.0

(e) n=2k, t=0.13s, c=57.4  (f) n=3.5k, t=0.14s, c=56.0  (g) n=16k, t=0.77s, c=53.7  (h) n=853k, t=55.5s, c=51.0

Figure 3: Performance of P-RRT* (a-d) and RRT* (e-h) in Local Minima Environment

**Lemma 3** *Since the expected limit of number of steps executed by RRT* at each iteration is atleast of the order of $\log(\mathcal{N}_n)$ time, implementing the computationally optimal algorithm given in [1] under fixed dimensions for computing nearest obstacle configuration, implies that there exists a constant $\phi_{RRT^*} \in \mathbb{R}_+$ i.e.,*

$$\lim_{n\to\infty} \mathbb{E}\left[\frac{\mathcal{S}_n^{\text{RRT}^*}}{\log(\mathcal{N}_n)}\right] \geq \phi_{\text{RRT}^*}$$

.

Hence, if lemma 3 holds, then it can be concluded that RRT* and P-RRT* has same asymptotic computational complexity as formalized in the following theorem.

**Theorem 9** *Assuming that lemma 3 holds, there exists a constant $\phi \in \mathbb{R}_+$ such that $\lim_{n\to\infty} \mathbb{E}\left[\frac{\mathcal{S}_n^{\text{P-RRT}^*}}{\mathcal{S}_n^{\text{RRT}^*}}\right] \leq \phi$ .*

# 7 Experimental Results

This section presents simulations performed on a 2.4GHz Intel corei5 processor with 4GB RAM. Performance results of our P-RRT* algorithm are compared with RRT*. For proper comparison, experimental parameters and configuration space size were kept same for both algorithms. Since sampling based algorithms exhibit large variations in results, the algorithms were run upto 50 times for each type of environment. Maximum, minimum and average number of iterations $n$ as well as time $t$ (in seconds) utilized by each algorithm to reach optimal path solution is presented in the table 1. To restrain the computa-

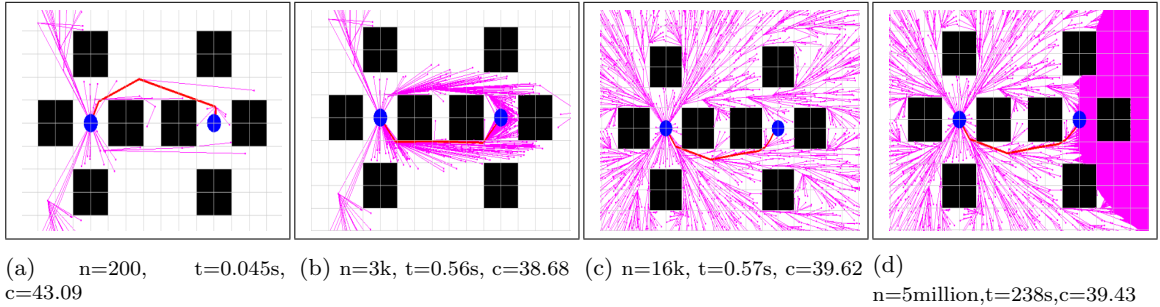

(a) n=200, t=0.045s, c=43.09    (b) n=3k, t=0.56s, c=38.68    (c) n=16k, t=0.57s, c=39.62    (d) n=5million, t=238s, c=39.43

Figure 4: Performance of P-RRT* (a-b) and RRT* (c-d) in 2D Cluttered Environment

tional time within reasonable limits, maximum limit for the number of tree nodes was kept at 5 million. The column fail in the tables denotes the number of runs for which the corresponding algorithm failed to find an optimal path solution within node limits. The variable $c^*$ represents the cost of optimal path returned by the algorithm in terms of Euclidean distance function. Moreover, for P-RRT* algorithm $\lambda = d^*_{\text{obs}} = 0.1$, while $k = 90$. However, it was noticed from the experiments that suitable range for $k$ is 80-100. Although, Artificial Potential Field Algorithm suffered from local minima problem, but, since the proposed P-RRT* algorithm only considers the attractive potential gradient, therefore, the P-RRT* algorithm does not inherit this limitation. Figure 3 demonstrates the working of both algorithms in a local minima environment. Figures 3a to 3d and Figures 3e to 3h show convergence progress of P-RRT* and RRT* respectively. It can be seen from these figures that P-RRT* is converging more quickly as compared to RRT* and unlike RRT*, the tree maintained by P-RRT* is directionalized towards the goal region.

Figures 4 show the working of P-RRT* and RRT* in a 2D cluttered environment. Figure 4a and 4b demonstrates the initial path and final path solution of P-RRT*. Moreover, it can be seen that P-RRT* takes lesser time and iterations (n=200, t=0.045s) as compared to RRT* (n=16063, t= 0.57s) for finding the initial path. Similarly, P-RRT* takes a reasonable number of iterations and time (n=3000, t= 0.65s) to find the optimal trajectory whereas RRT* is not able to find an optimal solution even after 5 million iterations. Moreover, in Figure 4d, half of the region is fully covered by pink red color coating, which is due to the large number of edges resulting from large number of samples that were generated in five million iterations.

Figure 5 represents two different complex mazes in 2D environment. In figure 5a and 5b, the start and goal regions, while very close together, require a path solution that traverses the length of a maze and stretches away from the goal. Once again, P-RRT* takes fewer iterations and therefore less time to find optimal path solution as compared to RRT*, as summarized in the table 1. Yet another complex maze environment is presented in Figure 5c and 5d, P-RRT* finds initial and final optimal path much faster than RRT*. Moreover, path returned by RRT* even after 5 million iterations is not an optimal path solution. Figures 6, 7 and 8 depict different scenarios in three dimensional space. Their results are summarized in the table I. It can be seen that a similar trend is followed by the algorithms in all environments i.e., P-RRT* rapidly converges to optimal as compared to RRT*.

Figure 10a compares the convergence rate of P-RRT* and RRT* in fifty different environments comprising of both 2D as well as 3D environments. Let initial feasible path denoted $\tau_{\text{init}} \in \sum_{\text{feasible}}$ is computed in $t_{\text{init}}$ time while optimal path solution denoted as $\tau^* \in \sum_{\text{feasible}}$ is computed in $t_{\text{opt}}$ time. Then the convergence rate is defined as $\dfrac{c(\tau_{\text{init}}) - c(\tau^*)}{t_{\text{opt}} - t_{\text{init}}}$. Since

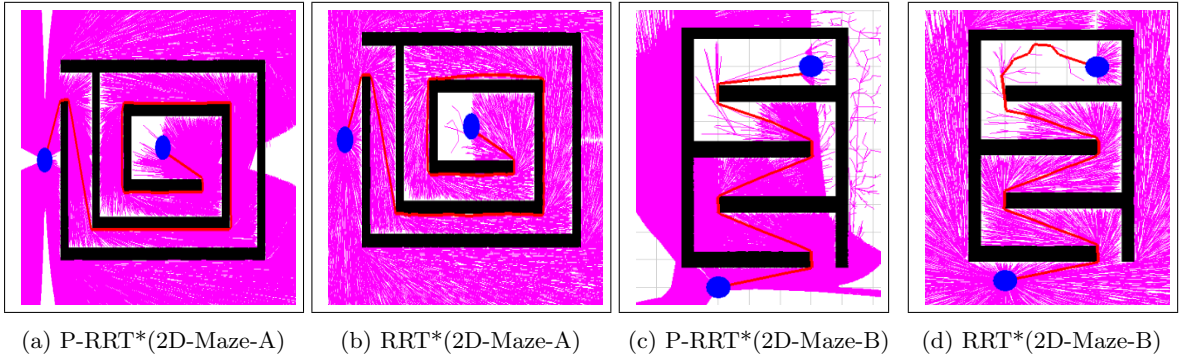

(a) P-RRT*(2D-Maze-A)    (b) RRT*(2D-Maze-A)    (c) P-RRT*(2D-Maze-B)    (d) RRT*(2D-Maze-B)

Figure 5: Performance of RRT* and P-RRT* in Complex Environment

the process of convergence to optimal path solution begins after finding initial feasible path solution, convergence rate is calculated after initial path computation. It can be seen in the figure 10a that the convergence rate of P-RRT* remains significantly higher than RRT*, which authenticates fast converging capability of P-RRT*.

Figure 10c shows memory consumed in bytes by these two algorithms to achieve optimal/near optimal path in twenty different environments. Since, P-RRT* uses lesser iterations as compared to RRT*, it therefore consumes lesser memory for any given environment.

Fixed cost is defined in terms of the average Euclidean distance of the most optimized path found by P-RRT* after several runs in a certain 3D environment. RRT* is tested to achieve this fixed cost in the same particular environment, and the result is shown in Figure 13. It can be seen from this figure that RRT* takes more time for converging the feasible path with *strong δ-clearance* to the feasible path with *weak δ-clearance* as compared to P-RRT*. Since RRT* consumes more time for converging the path solution therefore it has slower convergence rate as compared to P-RRT*.

Figure 14 shows the running time ratio of P-RRT* over RRT* after each iteration is executed. It can be seen that as the number of iterations increases, the running time ratio reaches a constant value. As a matter of fact, in this specific environment, the average amount of time taken by our proposed P-RRT* algorithm to determine a viable path to the goal was seen to be barely 1.6 times that of RRT*. Figure 10 demonstrates the effect of $k$ parameter, of P-RRT*, on exploration and exploitation of configuration space. It can be seen that the lower value of $k$ (figure 10a) biases the P-RRT* toward exploration while higher value (figure 10c) leads to more exploitation. It should be noted that the balance between exploitation and exploration is important to allow the algorithm to work in all types of environments. Figure 12 shows the working of P-RRT* and RRT*, under non-holonomic differential constrains, in a 2-D local minima environment. Figure 12a and 12b show the final path solutions of P-RRT* and RRT* respectively. Last row of table 1 summarizes the results of both the algorithm in this environment with differential constraints and it can be seen that P-RRT* takes lesser time and iterations (n=10431, t=1.93s) as compared to RRT* (n=954827, t= 62.9s) for finding the optimal path. Finally, figure 11 shows the implementation of P-RRT* on Poineer 3-DX robot using Player/Stage open source platform.

# 8 Conclusions and Future work

Recently, probabilistically complete sampling based motion planning algorithms have gained esteem due to their ability in finding a path irrespective of obstacles' geometry. RRT* assures asymptotic optimality but is not a memory efficient algorithm and has a

| Environment | Algorithm | $n_{\min}$ | $n_{\max}$ | $n_{\text{avg}}$ | $t_{\min}$ | $t_{\max}$ | $t_{\text{avg}}$ | $c^*$ | Fail |
|---|---|---|---|---|---|---|---|---|---|
| 2D-Local Minima (figures 3) | P-RRT* | 9261 | 10253 | 9582 | 1.73 | 1.92 | 1.85 | 51.0 | 0 |
| | RRT* | 851206 | 856121 | 853781 | 55.3 | 55.9 | 55.5 | 51.0 | 0 |
| 2D-Cluttered (figures 4) | P-RRT* | 2874 | 3411 | 3042 | 0.52 | 0.66 | 0.56 | 38.7 | 0 |
| | RRT* | - | - | - | - | - | - | - | 50 |
| 2D-Maze (A) (figures 5) | P-RRT* | 150686 | 152782 | 151178 | 29.2 | 28.9 | 29.1 | 163.1 | 0 |
| | RRT* | 4005814 | 4008126 | 4007931 | 259 | 260 | 260 | 163.2 | 7 |
| 2D-Maze (B) (figures 5) | P-RRT* | 254714 | 256921 | 254982 | 49.2 | 49.7 | 49.5 | 93.0 | 9 |
| | RRT* | - | - | - | - | - | - | - | 50 |
| 3D-Narrow Passages(figures 7) | P-RRT* | 46419 | 48726 | 47981 | 8.92 | 9.48 | 9.2 | 69.7 | 0 |
| | RRT* | 163319 | 168748 | 165261 | 10.9 | 11.6 | 11.2 | 69.9 | 0 |
| 3D-Multiple Barriers(figures 6) | P-RRT* | 84528 | 91827 | 87496 | 16.3 | 17.6 | 16.8 | 80.6 | 2 |
| | RRT* | 1941263 | 1978796 | 1961825 | 127 | 129 | 127 | 80.9 | 11 |
| 3D-Maze(figures 8) | P-RRT* | 41380 | 43861 | 42931 | 8.03 | 8.27 | 8.32 | 225.2 | 0 |
| | RRT* | 843428 | 849692 | 846452 | 54.9 | 55.6 | 55.2 | 226.1 | 4 |
| Poineer 3-DX robot(figures 12) | P-RRT* | 10126 | 11371 | 10431 | 1.91 | 2.13 | 1.93 | 61.2 | 0 |
| | RRT* | 951672 | 959165 | 954827 | 62.3 | 63.4 | 62.9 | 61.3 | 0 |

Table 1: Experimental results for computing optimal/near-optimal path.

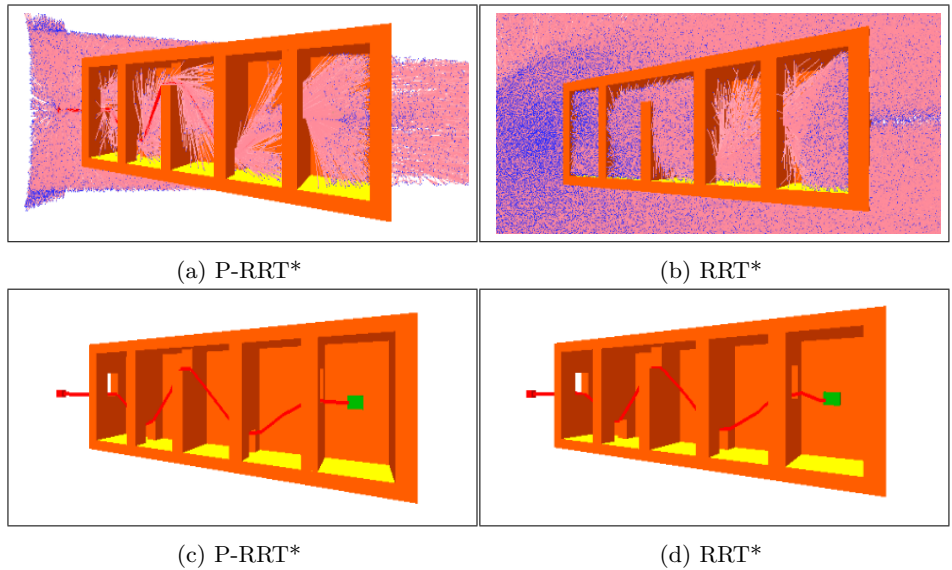

(a) P-RRT*  (b) RRT*

(c) P-RRT*  (d) RRT*

Figure 6: Performance of RRT* and P-RRT* in 3D environment with multiple barriers.

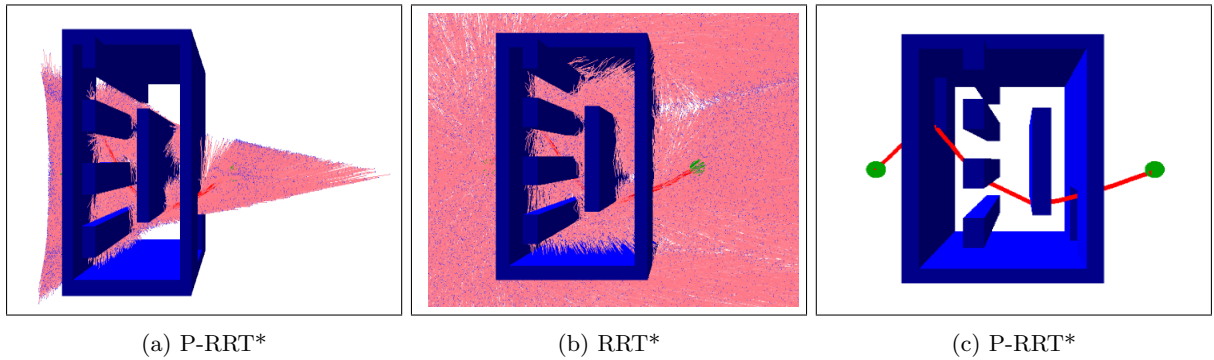

Figure 7: Performance of RRT* and P-RRT* in 3D environment with multiple narrow passages.

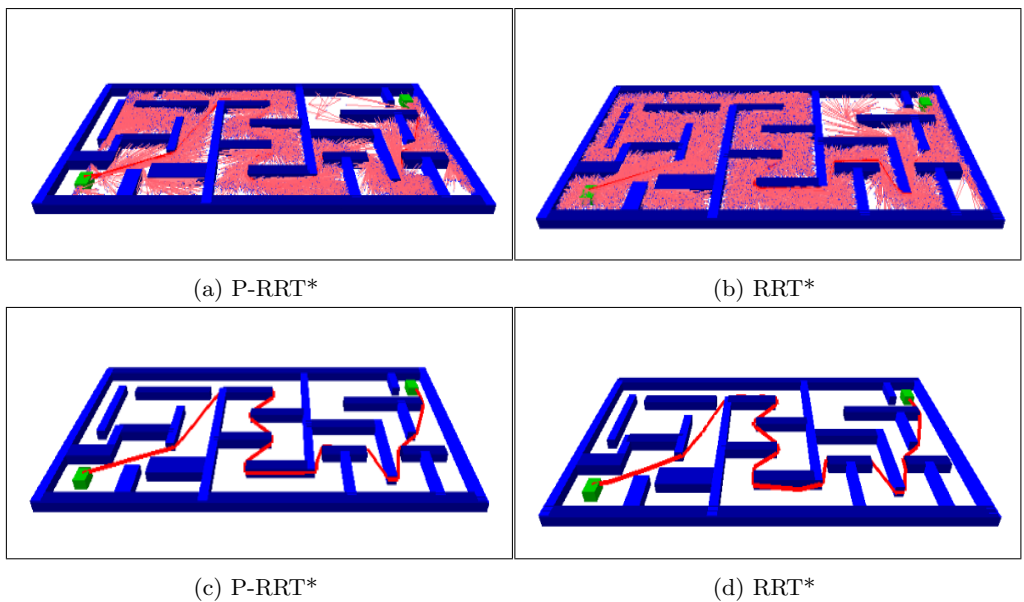

Figure 8: Performance of RRT* and P-RRT* in 3D complex maze.

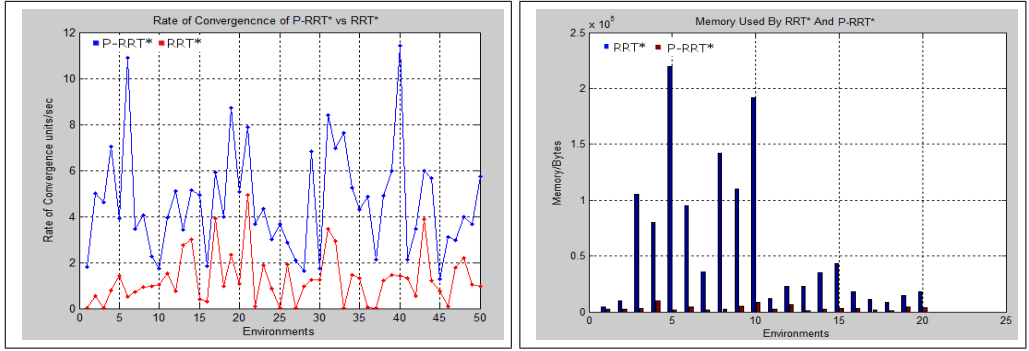

(a) Comparison of P-RRT* and RRT* in term of convergence rates.
(b) Memory Used by P-RRT* and RRT* in twenty different environments.

Figure 9: Performance of RRT* and P-RRT*.

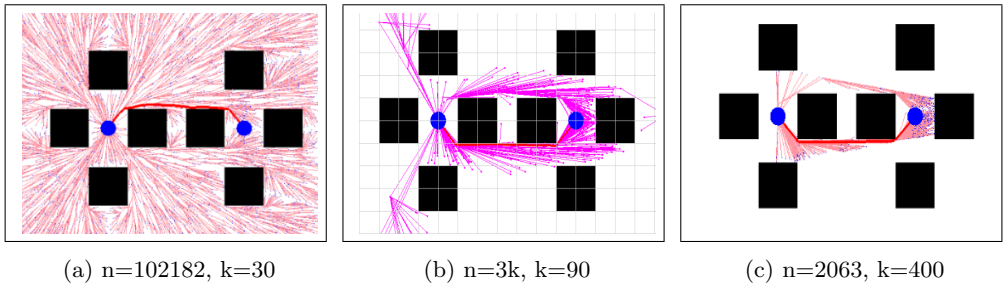

(a) n=102182, k=30  (b) n=3k, k=90  (c) n=2063, k=400

Figure 10: Visual representation of the role of $k$ on exploitation/exploration of configuration space.

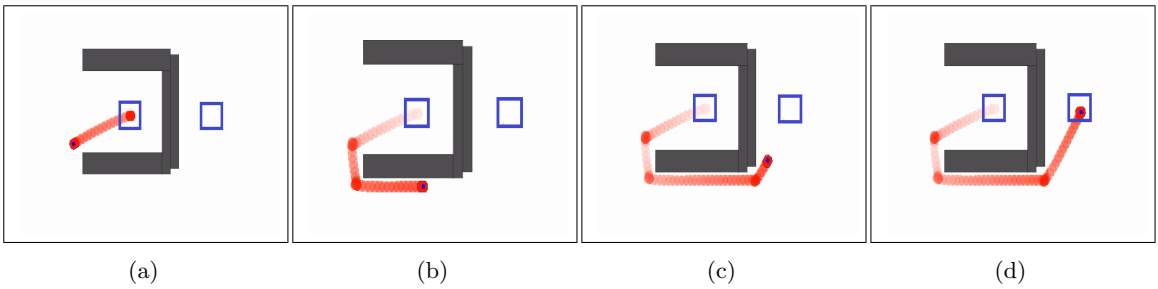

(a)  (b)  (c)  (d)

Figure 11: Demonstration of P-RRT* on non-holonomic Poineer 3-Dx robot using Player/Stage open source platform.

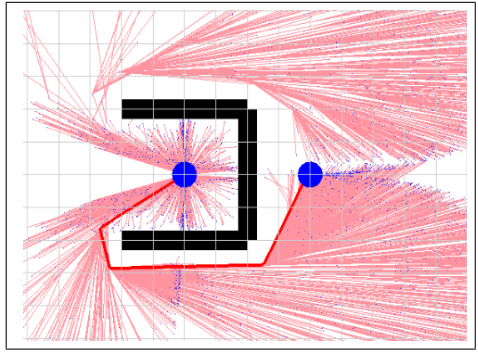

(a) P-RRT*: n=10431, t=1.93s

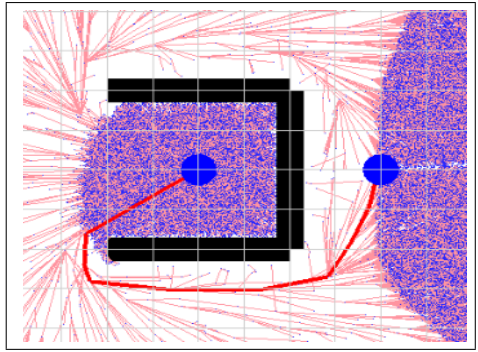

(b) RRT*: n=954827, t=62.9s

Figure 12: Optimal/near-optimal path computed by P-RRT* and RRT* in local minima environment under non-holonomic differential constraints.

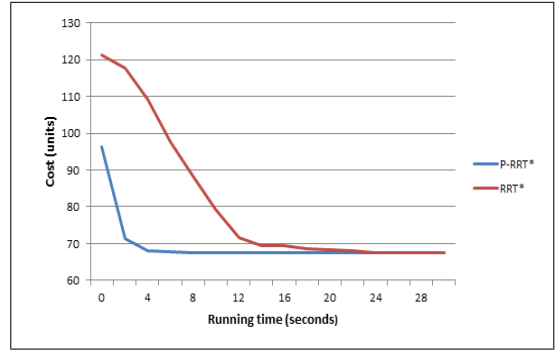

Figure 13: Cost vs running time of P-RRT* and RRT*.

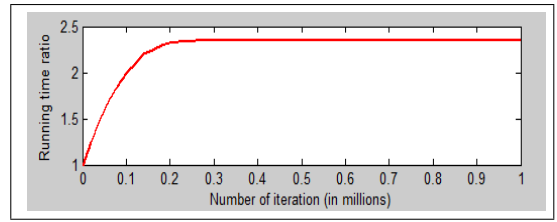

Figure 14: Running time ratio of P-RRT* over RRT*.

slow rate of convergence. Potential Function Based-RRT* (P-RRT*) addresses this problem and provides a solution by incorporating Artificial Potential Field Algorithm into RRT*. It is proven both experimentally and analytically that our proposed P-RRT* algorithm i) has same asymptotic computational complexity as that of RRT*; ii) inherits asymptotic optimality from RRT*; iii) does not suffer from local minima problem; iv) provides faster convergence to optimal path solution as compared to RRT*; v) utilizes lesser memory by sufficiently reducing number of iterations required and time consumed to compute a more optimized solution as compared to RRT*. In our future proceedings, we hope to employ P-RRT* for online motion planning, since the proposed algorithm allows faster convergence and determines the optimal path solution very quickly, therefore, it can be a very efficient solution to real time motion planning problems.